%% file: ms.tex
\documentclass[runningheads,hidelinks]{llncs}
\usepackage[utf8]{inputenc}
\usepackage{graphicx}
\usepackage{amsmath,amssymb} 
\usepackage{color}

\usepackage{tabularx}
\usepackage{booktabs}
\usepackage{multirow,array}
\usepackage{wrapfig}
\usepackage{import}
\usepackage{breqn}
\usepackage{xcolor}
\usepackage{booktabs}
\usepackage{fnpos}
\usepackage{xargs}		
\usepackage{tabularx}
\usepackage{pdflscape}
\usepackage{xspace}

\usepackage{amsmath}
\usepackage{amssymb}
\usepackage{url}
\usepackage{graphicx,subfigure}
\usepackage{multirow,array}
\usepackage{xcolor}
\usepackage{moresize}
\usepackage{booktabs}
\usepackage{nicefrac}

\usepackage{hyperref} 

\usepackage[normalem]{ulem}

\newcommand{\nyu}{NYUv2\xspace}
\newcommand{\dnet}{D3-Net\xspace}

\newcommand{\fnum}{$N$}

\def\ie{\emph{i.e.}\xspace}
\def\eg{\emph{e.g.}\xspace}

\def\etal{\emph{et al.}\xspace}

\definecolor{myblue}{RGB}{219,48,122}

\begin{document}

\def\ECCV18SubNumber{6}  

\title{Deep Depth from Defocus:\\ how can defocus blur improve 3D estimation using dense neural networks?}

\titlerunning{Deep DFD: improving 3D estimation with dense CNNs and defocus blur}

\author{Marcela Carvalho\inst{1} \and
Bertrand Le Saux\inst{1} \and
Pauline Trouv{\'e}-Peloux\inst{1} \and Andrés Almansa\inst{2} \and Frédéric Champagnat\inst{2}}

\authorrunning{Carvalho, Le Saux, Trouvé-Peloux, Almansa and Champagnat}

\institute{DTIS, ONERA, Universit\'{e} Paris-Saclay, FR-91123 Palaiseau, France
\email{\{name.lastname\}@onera.fr} \and
Universit\'{e} Paris Descartes, FR-75006 Paris, France\\
\email{\{name.lastname\}@parisdescartes.fr}}

\maketitle

\begin{abstract}
      \textit{Depth estimation is of critical interest for scene understanding and accurate 3D reconstruction. Most recent approaches in depth estimation with deep learning exploit geometrical structures of standard sharp images to predict corresponding depth maps. However, cameras can also produce images with defocus blur depending on the depth of the objects and camera settings. Hence, these features may represent an important hint for learning to predict depth. In this paper, we propose a full system for single-image depth prediction in the wild using  depth-from-defocus and neural networks. We carry out thorough experiments to test deep convolutional networks on real and simulated defocused images using a realistic model of blur variation with respect to depth. We also investigate the influence of blur on depth prediction observing model uncertainty with a Bayesian neural network approach. From these studies, we show that out-of-focus blur greatly improves the depth-prediction network performances. Furthermore, we transfer the ability learned on a synthetic, indoor dataset to real, indoor and outdoor images. For this purpose, we present a new dataset containing real all-focus and defocused images from a Digital Single-Lens Reflex (DSLR) camera, paired with ground truth depth maps obtained with an active 3D sensor for indoor scenes. The proposed approach is successfully validated on both this new dataset and standard ones as NYUv2 or Depth-in-the-Wild. Code and new datasets are available at \url{https://github.com/marcelampc/d3net_depth_estimation}.
      }
\end{abstract}

\section{Introduction}

3D reconstruction has a large field of applications such as in human computer interaction, augmented reality and robotics, which have driven research on the topic.  This reconstruction usually relies on on accurate depth estimates to process the 3D shape of an object or a scene.
Traditional depth estimation approaches exploit different physical aspects to extract 3D information from perception, such as stereoscopic vision, structure from motion, structured light and other depth cues in 2D images~\cite{saxena2009make3d,Calderero2013}. However, some of these techniques impose restrictions that depend on the environment (\eg sun, texture) or even require several devices (\eg camera, projector), leading to cumbersome systems. Many efforts have been made to build compact systems: the most notable are perhaps the light-field cameras which use a microlens array in front of the sensor, from which a depth map can be extracted~\cite{Ng2005}.

\begin{figure}[!t]
\centering
  \def\svgwidth{1.0\textwidth}
  \scalebox{1.0}{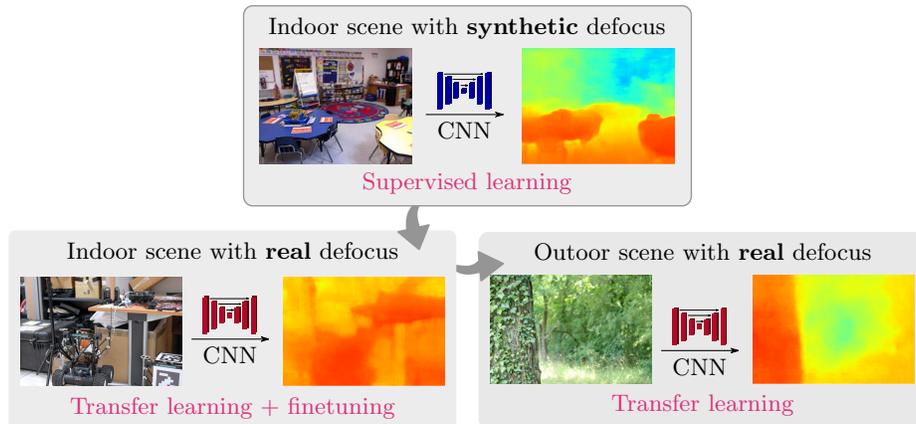}
  \caption{ Depth estimation with synthetic and real defocused data on indoor and outdoor challenging scenes. These results show the flexibility to new datasets of a model trained with a synthetically defocused indoor dataset, finetuned on a real DSLR indoor set and finally tested in outdoor scenes without further training. \label{fig:visual_abstract}}
\label{fig:catch}
\end{figure}

In recent years, several approaches for depth estimation based on deep learning, referred to as deep depth estimation, have been proposed, starting from \cite{Eigen2014}. These methods use a single point of view (a single image) and thus lead to compact, standard systems. Most of them exploit depth cues in the image based on geometrical aspects of the scene to estimate the 3D structure with the use of convolutional neural networks (CNNs)~\cite{Eigen2015,Li2015,Wang2015,Chakrabarti2016}. A few ones can also make use of additional depth cues such as stereo information to train the network~\cite{Ummenhofer2016} and improve predictions.

Another important cue for depth estimation has for long been the defocus blur. Indeed, Depth from Defocus (DFD) has been widely investigated in the past~\cite{Pentland87,Levin2007,Trouve13,martinello2011single,Sellent14,Zhuo11}. It led to various analytical methods and corresponding optical systems for depth prediction. However, DFD with a conventional camera and a single image suffers from ambiguity in depth estimation with respect to the focal plane and dead zone, due to the camera depth of field, where no blur can be measured. 
Moreover, DFD requires a scene model and an explicit calibration between blur level and depth value to estimate 3D information from an unknown scene.
It is tempting to integrate defocus blur with the power of neural networks, which leads to the question: does defocus blur improve deep depth estimation performances?

In this paper, we use a dense neural network, D3-Net~\cite{Carvalho2018icip}, in order to study the influence of defocus blur on depth estimation. 
Depth estimation performance is first tested on a synthetically defocused dataset created from \nyu, with optically realistic blur variation, which allows to compare several optical settings and study their influence. We further examine the uncertainty of the CNN predictions to better understand the main difficulties of the trained models while learning the proposed task with and without blur. We then propose to explore real defocused data with a new dataset which comprises of indoor all-in-focus and defocused images, and corresponding depth maps. Finally, we verify how the deep model behaves when confronted to challenging images in the wild with the Depth-in-the-Wild~\cite{chen2016wild} dataset and further outdoor images, with and without learning from defocus blur.

These experiments show that defocused information is exploited by neural networks and is indeed an important hint to improve deep depth estimation. Moreover, the joint use of structural and blur information proposed in this paper overcomes current limitations of single-image DFD such as ambiguity and dead zone, with respect to the focal plane. Finally, we show that these findings can be used in a dedicated device with real defocus blur to actually predict depth indoors and outdoors with good generalization.

\section{Related Work}

\textbf{Deep monocular depth estimation.}
 Several works have been developed to perform monocular depth estimation based on techniques of machine learning. One of the first solutions was presented by Saxena~\etal~\cite{Saxena2006}, which formulate the depth estimation for the Make3D dataset as a Markov Random Field~(MRF) problem with horizontally aligned images using a multi-scale architecture. 
More recent solutions are based on deep convolutional networks to exploit spatial correlation by enforcing a local connectivity through convolutional operations.  Eigen~\etal~\cite{Eigen2015,Eigen2014} proposed a multi-scale architecture capable of extracting global and local information from the scene to estimate the corresponding depth map. 
In~\cite{cao2017estimating}, Cao~\etal used a Conditional Random Field (CRF) to post-process the output of a deep residual network (ResNet)~\cite{he2016deep} in order to improve the reliability of the predictions. Xu~\etal~\cite{xu2017multi} adopted a deeply supervised approach connecting intermediate outputs of a ResNet to a continuous CRF fusion module to combine depth prediction at different scales achieving high performance for the task.
Also adopting residual connections, Laina~\etal~\cite{laina2016deeper} proposed an encoder-decoder architecture with fast up-projection blocks. More recently, Jung~\etal\cite{jung2017depth} introduced generative adversarial networks~\cite{Goodfellow} (GANs) to the deep depth estimation field, adapting an adversarial loss to refine the depth map predictions.
With a different strategy, 
\cite{Ummenhofer2016,Godard2016,Garg} propose to investigate the epipolar geometry using deep networks. DeMoN~\cite{Ummenhofer2016} jointly estimates a depth map and camera motion given a sequential pair of images exploring optical flow. The works in~\cite{Godard2016,Garg} use unsupervised learning to reconstruct stereo information and predict depth. More recently, Kendall and Gal~\cite{kendall2017uncertainties} and Carvalho~\etal~\cite{Carvalho2018icip} explore the reuse of feature maps during learning, building upon an encoder decoder with dense and skip connections~\cite{huang2017densenet}. In~\cite{kendall2017uncertainties}, they propose a regression function that captures the uncertainty of the data and in~\cite{Carvalho2018icip} the network adopts an end-to-end adversarial generative loss function to improve prediction.

The aforementioned techniques for monocular depth estimation with neural networks base their learning capabilities on structured information (\eg, textures, linear perspective, statistics of objects and their positions).
However, depth perception can use another well-know cue: defocus blur.
We present in the following section state-of-the-art approaches from this domain.

\textbf{Depth estimation using DFD.}
In computational photography, several works investigated the use of defocus blur to infer depth, starting from \cite{Pentland87}. Indeed, the amount of defocus blur of an object can be related to its depth using geometrical optics $\epsilon = Ds \cdot \left|{\frac{1}{f} - \frac{1}{d_{out}} - \frac{1}{s}}\right|$,
where $f$ stands for the focal length,
$d_{out}$ the distance of the object with respect to the lens, $s$ the distance between the sensor and the lens and $D$ the lens diameter. $D=\nicefrac{f}{N}$, where $N$ is the f-number.

\begin{wrapfigure}{r}{60mm}
  \vspace{-5mm}
  \centering
  \resizebox{0.4\textwidth}{!}{\includegraphics{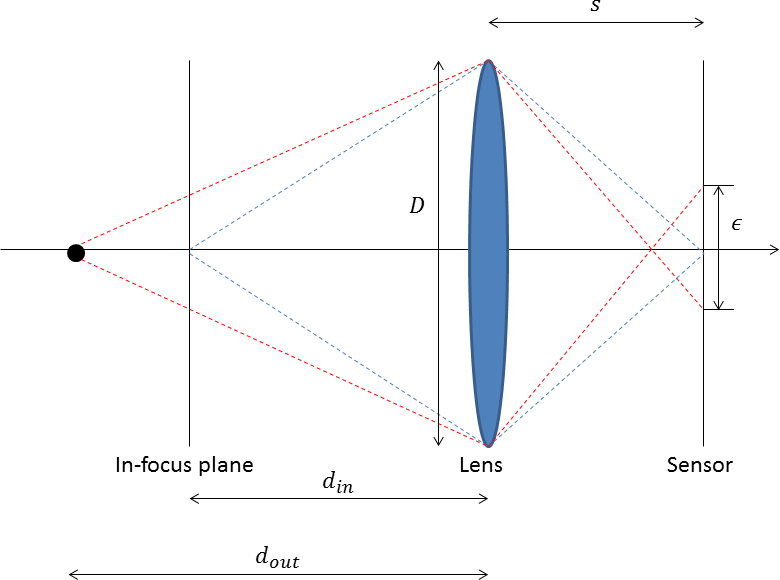}}
  \caption{Illustration of the DFD principle.  Rays originating from the out of focus point (black dot) converge before the sensor and spread over a disc of diameter $\epsilon$.}
  \label{fig:DFD_principle}
\end{wrapfigure}
Recent works usually use DFD with a single image (SIDFD). Although the acquisition is simple, it also leads to more complex processing as both the scene and the blur are unknown. State of the art approaches use analytical models for the scene such as sharp edges models \cite{Zhuo11} or statistical scene Gaussian priors \cite{levin2009understanding,Trouve13}.
Coded apertures have also been proposed to improve depth estimation accuracy with respect to standard optics~\cite{Levin2007,veeraraghavan2007dappled,Chakrabarti12,Sellent14}.

Nevertheless, SIDFD suffers from two main limitations: first, there is an \textit{ambiguity} related to the object's position in front or behind the in-focus plane; second, blur variation cannot be measured in the camera depth of field, leading to a \textit{dead zone}. Ambiguity can be solved using asymmetrical coded aperture \cite{Sellent14}, or even by setting the focus at infinity, at a cost of reducing the light intensity that reaches the sensor or large depth of field (\ie, dead zone), respectively. Second, dead zones can be overcome using several images with various in-focus planes. 
In a single snapshot context, this can be obtained with unconventional optics such as a plenoptic camera~\cite{hazirbas17ddff} or a lens with chromatic aberration~\cite{Guichard2009,Trouve13}, but both at the cost of image quality (low resolution or chromatic aberration).

Indeed, inferring depth from the amount of defocus blur with model-based techniques requires a tedious explicit calibration step, usually conducted using point sources or a known high frequency pattern \cite{Delbracio2011,Levin2007} at each potential depth.
These constraints lead us to investigate data-based methods using deep learning techniques to explore structured information together with blur cues to execute the proposed task.\\

\textbf{Learning depth from defocus blur.}
The existence of common datasets for depth estimation~\cite{Silberman2012NYUv2,saxena2009make3d,hazirbas17ddff}, containing pairs of RGB images and corresponding depth maps, facilitates the creation of synthetic defocused images using real camera parameters. Hence, a deep learning approach can be used.
To the best of our knowledge, only a few papers in the literature use defocus blur as a cue in learning depth from a single image. Srinivasan~\textit{et al.}.~\cite{srinivasan2017} uses defocus blur to train a network dedicated to monocular depth estimation: the model measures the consistency of simulated defocused images, generated from the estimated depth map and all-in-focus image, with true defocused images. However, the final network is used to conduct depth estimation from all-in-focus images. Hazirbas~\etal~\cite{hazirbas17ddff} propose to conduct depth estimation using a focal stack, which is more related to depth from focus approaches than DFD. Finally, \cite{anwardepthblur} presents a network for depth estimation and deblurring using a single defocused image.
This work shows that networks can integrate blur interpretation. However, ~\cite{anwardepthblur} creates a synthetically defocused dataset from real NYUv2 images without consideration of a realistic blur variation with respect to the depth, nor sensor settings (\eg, camera aperture, focal distance). 
However, there has not been much investigation about how defocus blur influence on depth estimation, nor how can these experiments improve depth prediction in the wild.  \\

In contrast with previous works, to the best of our knowledge, 
we present the first system for deep depth from defocus (Deep-DFD): \ie single-image depth prediction in the wild using deep learning and depth-from-defocus.
In section~\ref{sec:nyuv2_exp}, we study the influence of defocus blur on deep depth estimation performances.
(i) We run tests on a synthetically defocused dataset generated from a set of true depth maps and all-in-focus images. The amount of defocus blur with respect to depth varies according to a physical optical model to better relate to realistic examples. (ii) We also compare performances of deep depth estimation with several optical settings: here we compare the case of all-in-focus images with the case of defocused images from three different focus settings. (iii) We analyse the influence of defocus blur on neural networks using uncertainty maps and diagrams of errors per depth. 
In section~\ref{sec:real_dfd}, (iv) we carry out validation and analysis of the estimation results on a new dataset with pairs of real images and depth maps obtained with a Digital Single Lens Reflex (DSLR) camera and an RGB-D (Red Green Blue Depth) sensor. At last, in section~\ref{sec:diw}, (v) we show the network
is able to generalized to images in the wild.

\section{Learning DFD to Improve Depth Estimation\label{sec:nyuv2_exp}}
In this section, we perform a series of experiments with synthetic and real defocused data exploring the power of deep learning to depth prediction.  
As we are interested in using blur as a cue, we do not apply any image processing for data augmentation capable of modifying out-of-focus information. Hence, for all experiments, we extract random crops of 224x224 from the original images and apply horizontal flip with a probability of 50\%. Tests are realized using the full-resolution image.

\subsection{D3-Net Architecture}
\begin{figure}
  \centering
  \includegraphics[width=0.95\linewidth]{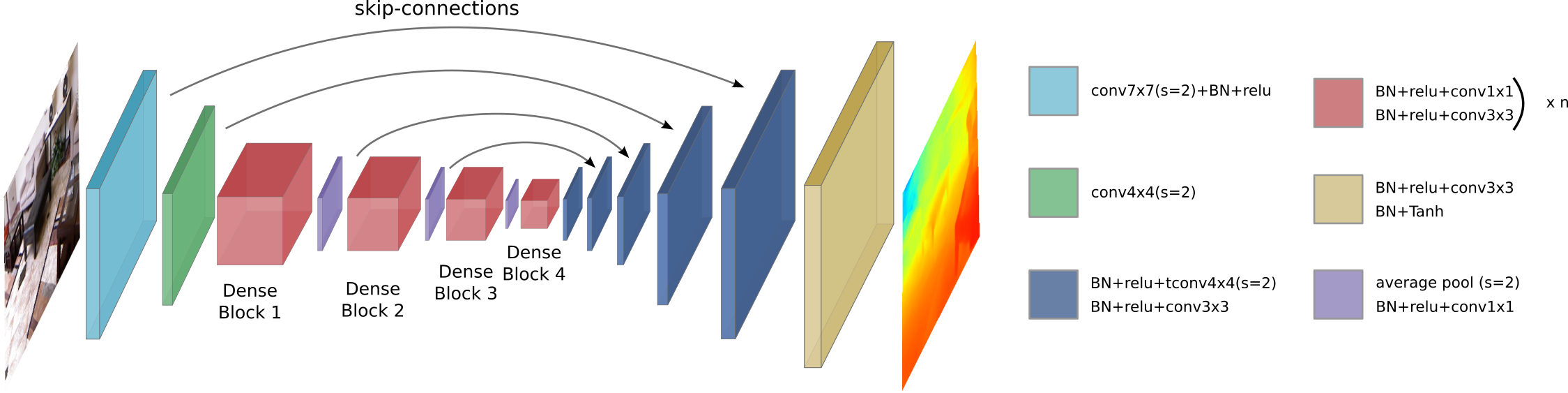}
  \caption{\label{fig:d3net_architecture} D3-Net architecture from~\cite{Carvalho2018icip}. The encoder part corresponds to a DenseNet-121~\cite{huang2017densenet}, with $n=6,12,24,16$, respectively for indicated Dense Blocks. The encoder-decoder structure is based on U-Net~\cite{Ronneberger2015} to explore the reuse of feature maps.}
\end{figure}
To perform such tests, we adopt the D3-Net architecture from~\cite{Carvalho2018icip}, illustrated in figure~\ref{fig:d3net_architecture}.
We use the PyTorch framework on a NVIDIA TITAN X GPU with 12GB of memory. We initialize the D3-Net encoder, corresponding to DenseNet-121, with pretrained weights on Imagenet dataset and D3-Net decoder with random weights from a normal distribution with zero-mean and 0.2 variance. We add dropout~\cite{srivastava2014dropout} regularization with a probability of 0.5 to the first four convolutional layers of the decoder as we noticed it improves generalization. We also adopt dropout layers to posteriorly study model's uncertainty.

\subsection{Synthetic NYUv2 with defocus blur} 
The NYU-Depth V2 (NYUv2) dataset~\cite{Silberman2012NYUv2} has approximately 230k pairs of images from 249 scenes for training and 215 scenes for testing. In \cite{Carvalho2018icip}, D3-Net reaches its best performances when trained with the complete dataset. However, \nyu also contains a smaller split with 1449 pairs of aligned RGB and depth images, of which 795 pairs are used for training and 654 pairs for testing. Therefore, experiments in this section were performed using this smaller dataset 
to fasten experiments. Original frames from Microsoft Kinect output have the resolution of 640x480. Pairs of images from the RGB and Depth sensors are posteriorly aligned, cropped and processed to fill-in invalid depth values. Final resolution is 561x427.

\begin{wrapfigure}{l}{55mm}
  \centering
  \vspace{-5mm}
  \includegraphics[width=58mm]{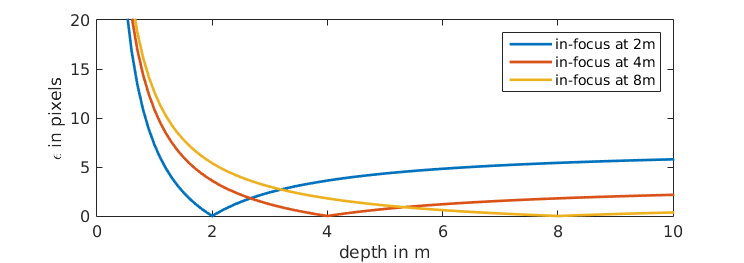}
  \caption{Blur diameter variation vs depth for the following in-focus settings: 2m, 4m and 8m tests on the NYUv2 dataset.}
  \label{fig:var_blur}
  \vspace{-5mm}
  \end{wrapfigure}

\begin{figure}[h]
  \centering
  \def\svgwidth{1.0\textwidth}
  \scalebox{1.0}{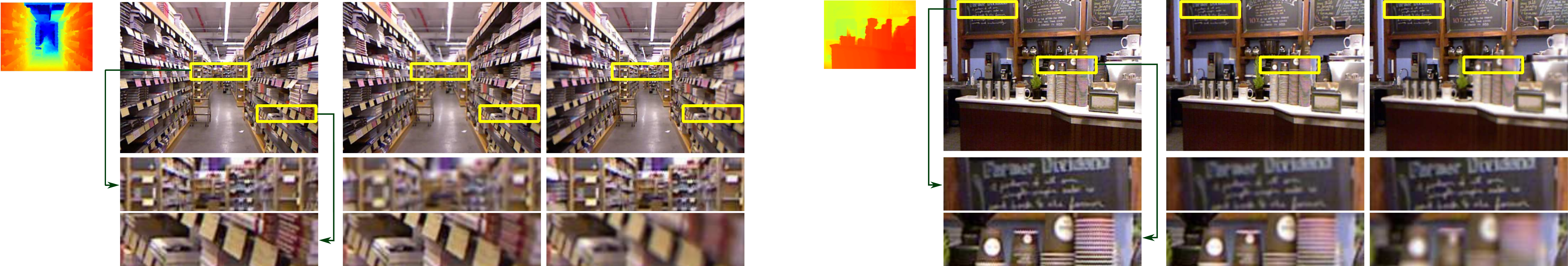}
  \caption{Examples of synthetic defocused images generated from an image of the NYUv2 database for two camera in-focus plane settings: 2 and 8 m.}
  \label{fig:visu_synthetic_images_NYUV2}
\end{figure}

To generate physically realistic out-of-focus images, we choose the parameters corresponding to a synthetic camera with a focal length of 15mm, f-number 2.8 and pixel size of 5.6$\mu$m. Three settings of in-focus plane are tested, respectively at 2m, 4m and 8m from the camera. Figure~\ref{fig:var_blur} shows the variation of the blur diameter $\epsilon$ with respect to depth, for both settings and Figure~\ref{fig:visu_synthetic_images_NYUV2} shows examples of synthetic defocused images.
As illustrated in Figure~\ref{fig:var_blur}, setting the in-focus plane at 2m corresponds to a camera with small depth of field. The objects in the depth range of 1 to 10m will present small defocus blur amounts, apart from the objects in the camera depth of field, which remain sharp. Note that this configuration suffers from depth ambiguity caused by the blur estimation. Setting the in-focus plane at a larger depth, here 4m or 8m, corresponds to a camera with larger depth of field. Only the closest objects will show defocus blur, with the blur amount in the approximate depth range 0-3m that will be larger than with the 2m setting. This can be observed in the extracted details of images in Figure~\ref{fig:visu_synthetic_images_NYUV2}.

To create the out-of-focus dataset, we adopt the layered approach of~\cite{Hasinoff2007} where each defocused image $\widehat{L}$ is the sum of $K$ blurred images multiplied by masks taking into account local object depth and occlusion of foreground objects according to:
\begin{equation}
\widehat{L}=\sum_k \left[(A_kL+A^*_{k}L^*_{k})* h(k))\right] M_k,
\end{equation}
where $h(k)$ is the defocus blur at depth $k$, $L$ is the all-in-focus image, $A_k$ is the mask corresponding to object at depth $k$ and $A^*_{k}L^*_{k}$ the layer extension behind occluders, obtained by inpainting. Finally $M_k$  models the cumulative occlusions defined as: 
\begin{equation}
M_k=\prod_{k'=k+1}^{K}(1-A_{k'}*h(k')).
\end{equation}
Following \cite{srinivasan2017}, we chose to model the blur as a disk function of which the diameter varies with the depth.

As will be discussed later in this paper, the proposed approach can be disputable as the true depth map is used to generate the out-of focus image. 
However, this strategy allows us easily perform various experiments to analyze the influence of blur corresponding to different in-focus settings in the image.

\subsection{Performance results}

Table~\ref{table:perfo_comp_nyuv2} shows performance results of \dnet first using all-focuses and then defocused images with proposed settings. Note that as illustrated in Figure~\ref{fig:var_blur} when  the in-focus plane is at 8m, there is no observable ambiguity. Hence performance comparison with SIDFD methods can then be made. In such manner, we include error metrics of two methods from the SIDFD literature~\cite{Zhuo11,Trouve11} which estimate the amount of local blur using either sharp edge model or gaussian prior on the scene gradients.

\begin{figure}[t]
  \centering
  \def\svgwidth{0.9\textwidth}
  \scalebox{1.0}{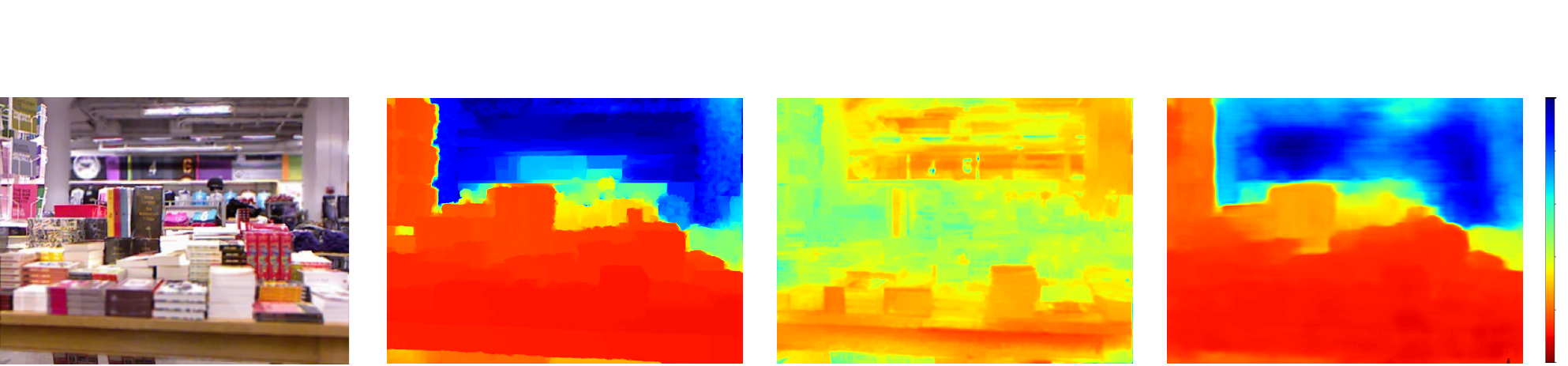}
  \caption{Comparison between D3-Net estimation and Zhuo~\cite{Zhuo11} for images with the focus plane at 2m.}
  \label{fig:illu_ambi}
\end{figure}

Several conclusions can be drawn from Table~\ref{table:perfo_comp_nyuv2}. First, as already stated by Anwar~\etal, there is a significant improvement on the performance of depth estimation when using out-of-focus images instead of all-in-focus images. Second, D3-Net outperforms the standard model-based SIDFD methods, which can also be observed in figure~\ref{fig:illu_ambi}, without requiring an analytical scene model nor explicit blur calibration. Indeed, the neural network makes use of both parameters without being specifically designed to. Furthermore, there is also a sensitivity of the depth estimation performance with respect to the position of the in-focus plane. The best setting for these tests is the in-focus plane at 2m, which corresponds to a significant amount of blur for most of the objects but near the focal plane. This shows that the network actually uses blur cue and is able to overcome depth ambiguity using geometrical structural information. Figure \ref{fig:illu_ambi} also illustrates this conclusion: the presented scene has mainly three depth levels with a foreground below 2m, a background after 2m, and intermediate level around 2m. The corresponding out-of-focus image is generated using an in-focus plane at 2m. Using \cite{Zhuo11}, the background and the foreground are at the same depth, while D3-Net shows no such error in the depth map.

Finally, we also trained and tested D3-Net with the dataset proposed in Anwar~\etal~\cite{anwardepthblur}. However, differently from the method explored in this paper, the out-of-focus images were generate without any regard to camera settings. The last two lines from Table~\ref{table:perfo_comp_nyuv2} shows that D3-Net also outperforms the network in~\cite{anwardepthblur}. 

  \begin{table}[]
  \centering
  \caption{\label{table:perfo_comp_nyuv2} Performance comparison of D3-Net using all-in-focus images, defocused images with three positions of the in-focus planes, and two SIDFD approaches~\cite{Zhuo11,Trouve11} for the 8m focus setting. }
  \small
  \scalebox{1.0}{
  \begin{tabular}{p{3.7cm}@{\hspace{1.2mm}}c@{\hspace{1.2mm}}c@{\hspace{1.2mm}}c@{\hspace{0.5mm}}c@{}r@{}r@{}r@{\hspace{1mm}}r@{\hspace{1mm}}r@{}}\toprule
  {\multirow{2}{*}{Methods}} & \multicolumn{4}{c}{Error$\downarrow$} & \phantom{abc} &  \multicolumn{3}{c}{Accuracy$\uparrow$} \\
  \cmidrule{2-5} \cmidrule{7-9}
  & rel & log10 & rms & rmslog &&  \scalebox{.65}{$\delta\!<\!1.25$} & \scalebox{.65}{$\delta\!<\!1.25^2$} & \scalebox{.65}{$\delta\!<\!1.25^3$}\\  \midrule
  \multicolumn{9}{c}{Original RGB images 
  } \\ 
  \midrule
  D3-Net All-in-focus & 0.226 & - & 0.706 & - && 65.8\% & 89.2\% & 96.7\% \\
  \midrule
  \multicolumn{9}{c}{RGB images with additional blur 
  }\\ \midrule
  D3-Net 2m focus & 0.068 & 0.028 & 0.274 & 0.110 && 96.1\% & 99.0\% & 99.6\% \\
  D3-Net 4m focus & 0.085 & 0.036 & 0.398 & 0.125 && 92.5\% & 99.0\% & 99.8\% \\
  D3-Net 8m focus & 0.060 & - & 0.324 & - && 95.2\% & 99.1\% & 99.9\% \\
  {Zhuo \etal  \cite{Zhuo11} 8m focus} & 0.273 & - & 0.981 & - && 51.7\% & 83.1\% & 95.1\% \\
  Trouv\'e \etal~\cite{Trouve11} 8m focus & 0.429 & 0.289 & 1.743 & 0.956 && 39.2\% & 52.7\% & 61.5\%\\
  \midrule
  \multicolumn{9}{c}{RGB images with additional blur proposed by~\cite{anwardepthblur}} \\
  \midrule
  Anwar \etal~\cite{anwardepthblur} & 0.094 & 0.039 & 0.347 & - && - & - & - \\
  D3-Net & 0.036 & 0.016 & 0.144 & 0.054 && 99.3\% & 100.0\% & 100.0\% \\
  \bottomrule
  \end{tabular}
  }
  \vspace{-5mm}
\end{table}

\begin{figure}[h]
\centering
  \def\svgwidth{0.9\textwidth}
  \scalebox{1.0}{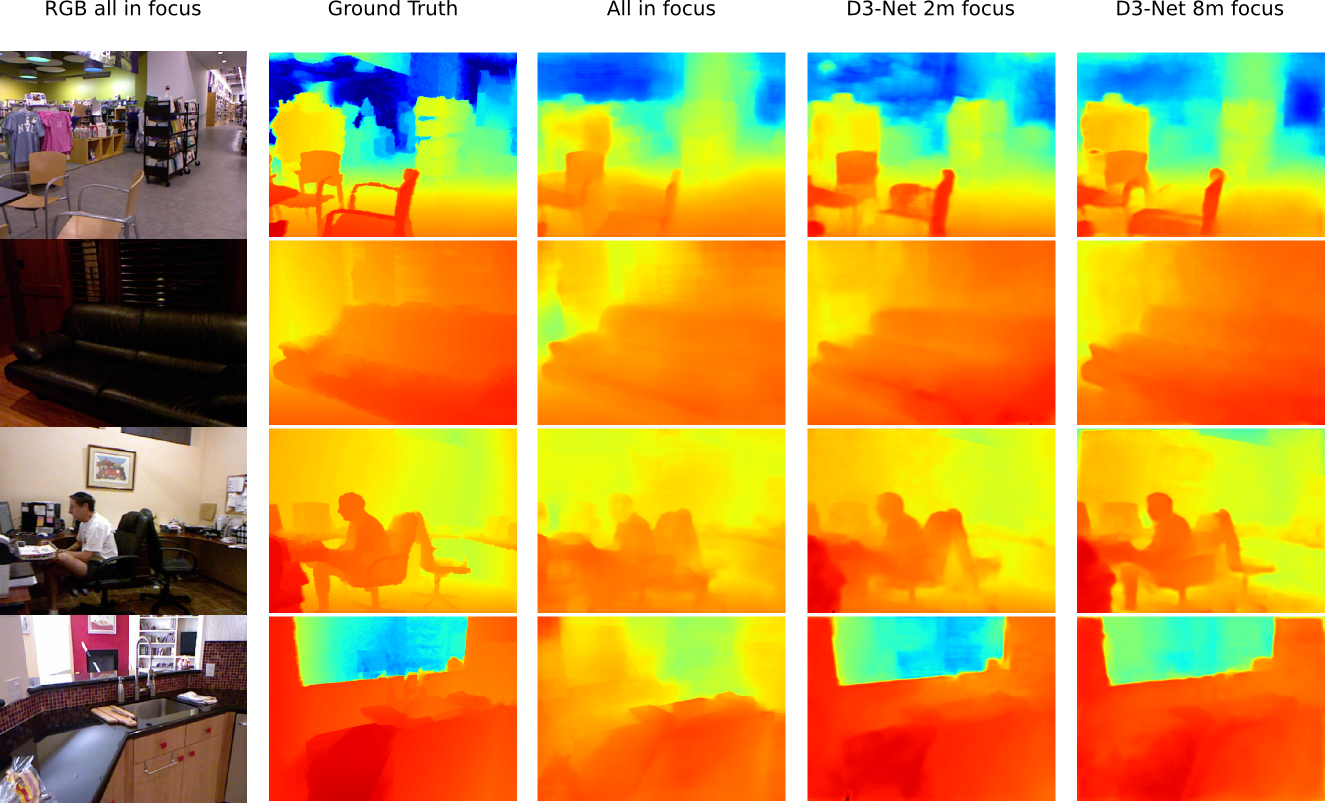}
  \caption{Qualitative comparison for different predictions with the proposed defocus blur configurations.}
\label{fig:nyu_prediction}
\end{figure}

In addition, Figure~\ref{fig:nyu_prediction} and columns 3 and 6 from Figure~\ref{fig:nyu_uncertainties} show examples of predicted depth maps. The depth maps obtained with out-of-focus images are sharper than using all-in-focus images. Indeed, defocus blur provides local depth information to the network leading to a better depth map segmentation. \\

\begin{wrapfigure}{l}{0.5\linewidth}
  \vspace{-5mm}
  \centering
  \includegraphics[width=0.95\linewidth]{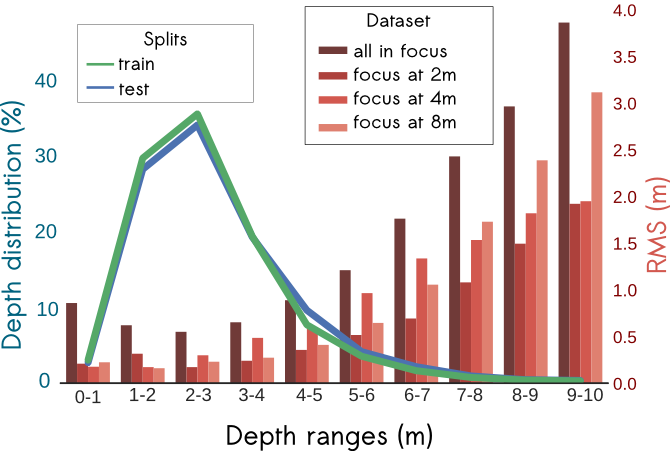}
  \caption{Distribution of depth pixels on different depth ranges and RMS performance of D3-Net trained on all-focused and defocused models.}
  \label{fig:dist_plotbar_nyuv2}
  \vspace{-5mm}
  \end{wrapfigure}

\textbf{Per depth error analysis.} There is an intrinsic relation between the number of examples a network can learn from and its performance when observing similar samples to them. Here, we study the prediction error per depth range when using all-in-focus images or defocused images and observe relation to depth data distribution. Figure~\ref{fig:dist_plotbar_nyuv2} shows in the same plot repartition the RMS per depth in meters and the depth distribution for testing and training images for the NYUv2 dataset.

  For all-in-focus images, the errors seem to be highly correlated to the number of examples in the dataset. Indeed, a minimum error is obtained for 2m, corresponding to the depth with the highest number of examples. On the other hand, using defocus blur, errors repartition is more similar to a quadratic increase of error with depth, which is the usual error repartition of passive depth estimation. 
  
Furthermore, the 2m focus setting does not show an error increase at 2m (its focal plane position), though it corresponds to the dead zone of SIDFD. This surprising result shows that the proposed approach overcomes this issue probably because the neural network relies on context and geometric features. In general, 2m, 4m and 8m focus have similar performance for depth range between 0 to 3m. After this depth, the 2m focus presents lowest errors. When focus is at 4m, we observe a drop in all metrics performances compared to 2m and 8m. The reason for this can be observed when comparing both Figures~\ref{fig:var_blur} and \ref{fig:dist_plotbar_nyuv2}. This configuration presents worst RMS performances between 3 and 7m, when blur information is too small to be used by the network and there is not enough data to overcome the missing cue, but enough to worsen results. The same happens to the model at 8m, where results are more prone to errors after approximately 7m. 

\subsection{Uncertainties on the Depth Estimation Model}

To go further in the analysis of understanding the influence of blur in depth prediction, we present a study on model uncertainties following~\cite{kendall2015bayesian,kendall2017uncertainties,gal2016dropout}. More precisely, we evaluate the epistemic uncertainty of the deep network model, or how ignorant is the model with respect to the dataset probabilistic distribution.  

To perform this experiment, we place a prior distribution over the network weights to replace the deterministic weight parameters at test time~\cite{kendall2017uncertainties}. We adopt the Monte Carlo dropout method~\cite{gal2016dropout} to measure variational inference placing dropout layers during train and also during test phases. Following~\cite{kendall2015bayesian}, we produce 50 samples for each image, calculate the mean prediction and the variance of these predictions to generate the model uncertainty.

\begin{figure}[]
  \centering
  \def\svgwidth{1.85\textwidth}
  \scalebox{0.55}{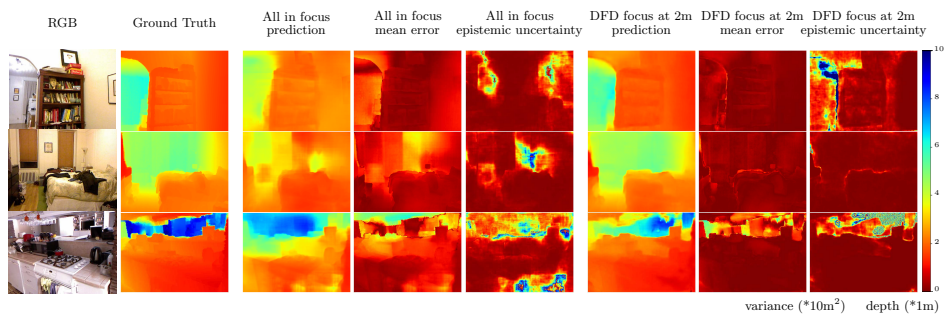}
  \caption{Qualitative comparison of all focus and DFD with 2m focus prediction, mean error and epistemic uncertainty with NYUv2 dataset. Lower values of depth and uncertainties are represented by warmer colors.}
  \label{fig:nyu_uncertainties}
  \vspace{-10mm}
\end{figure}

Figure~\ref{fig:nyu_uncertainties} presents examples of the network prediction, mean error and epistemic uncertainty for the \nyu dataset with sharp images and with focus at 2m. Mean error is produced using the ground truth image, while the variance only depends on the model's prior distribution. For both configurations, highest variances are observed in non-textured areas and edges, as predictable. However, the model with blur has less diffuse uncertainty: it is concentrated on the object edges, and these objects are better segmented. 
In the second row of the figure, we observe that the all-in-focus model has difficulties to find an object near the window, while this is overcome with blur cues present on the defocused model. In the first row, we observe high levels of uncertainty at the zones near the bookcase, defocused model reduce some of this variance with defocus information. Finally, the last row presents a hard example where both models have high prediction variances mainly in the top middle part, where there is a hole. However the all-in-focus model also presents high mean error and variance in the bottom zone unlike the model with blur.

\section{Experiments on a Real Defocused Dataset\label{sec:real_dfd}}

\begin{wrapfigure}{r}{30mm}
  \centering
  \vspace{-5mm}
  \includegraphics[width=.8\linewidth]{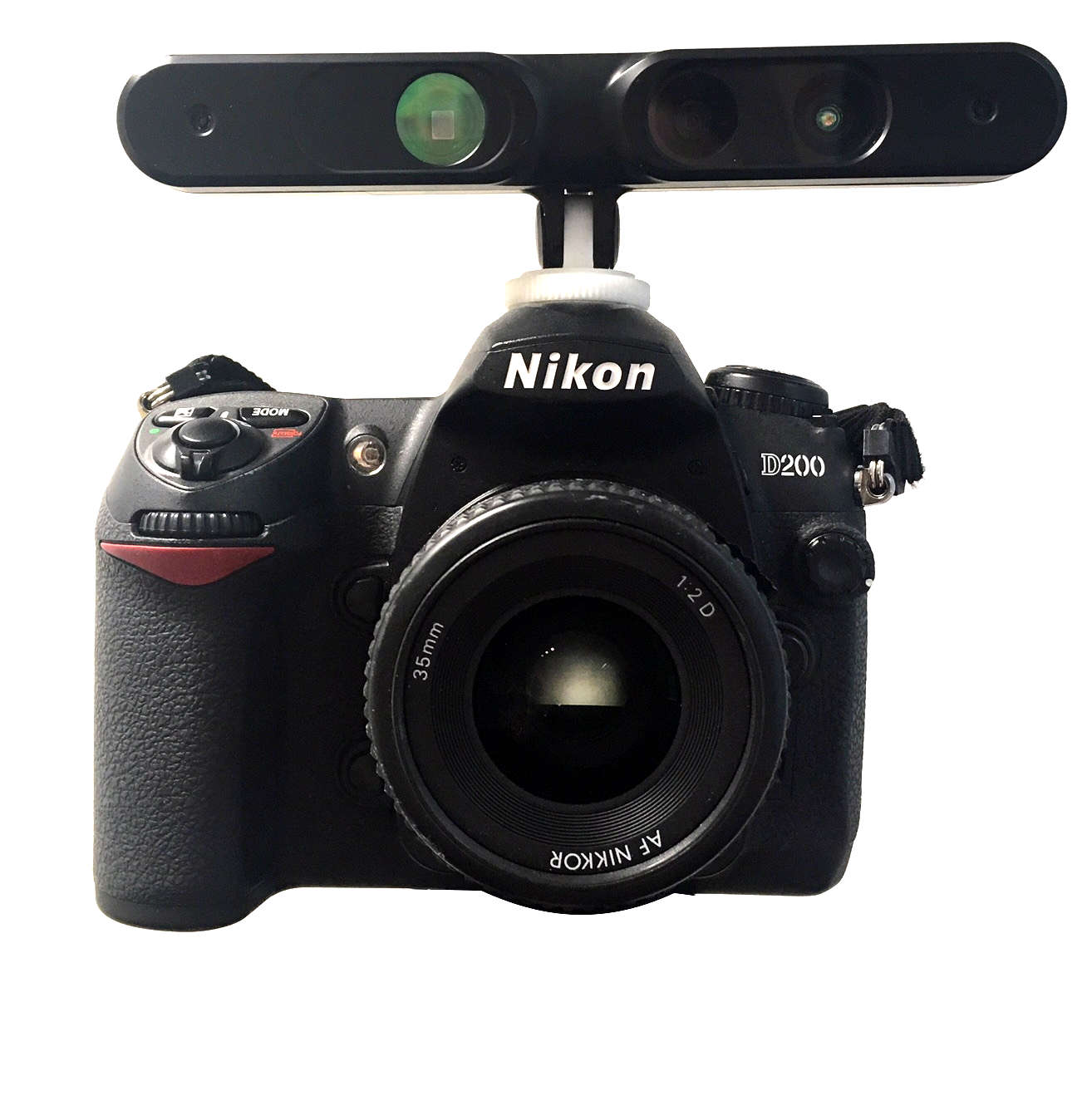}
  \caption{\label{fig:dslr_xtion} Experimental platform with Xtion PRO sensor coupled to a DSLR Nikon camera.}
\end{wrapfigure}
In section~\ref{sec:nyuv2_exp}, several experiments were performed using a synthetic version of \nyu. 
However, when adopting convolutional neural networks, it can be a little tricky to use the desired output (depth) to create blur information on the input of the network. So, in this section, we propose to validate our method 
on real defocused data from a DSLR camera paired with the respective depth map from a calibrated RGB-D sensor. 

\textbf{Dataset creation.} To create a DFD dataset, we paired a DSLR Nikon D200 with an Asus Xtion sensor to produce out-of-focus data and corresponding depth maps, respectively. Our platform can be observed in Figure~\ref{fig:dslr_xtion}. 
We carefully calibrate the depth sensor to the DSLR coordinates to produce RGB images paired with the corresponding depth map. The proposed dataset contains 110 images from indoor scenes, with 81 images for training and 29 images for testing. Each scene is acquired with two camera apertures: \fnum=2.8 and \fnum=8, providing respectively out-of-focus and all-in-focus images.

As the DFD dataset contains a small amount of images, we pretrain the network using simulated images from \nyu dataset and then conduct a finetuning of the network using the real dataset.  The DSLR camera originally captures images of high resolution 3872x2592; but to reduce the calculation burden, we downsample the DSLR images to 645x432. In order to simulate defocused images from \nyu as similar as possible as those provided by the DSLR, the image from the Kinect are upsampled and cropped to have the same resolution and the same field of view as the 645x432 DSLR images. Then defocus blur is applied to the images using the same method as in section \ref{sec:nyuv2_exp} but with a blur variation with that fits the real blur variation of the DSLR, obtained experimentally.

\begin{figure}[ht]
  \centering
  \def\svgwidth{0.9\textwidth}
  \scalebox{1.0}{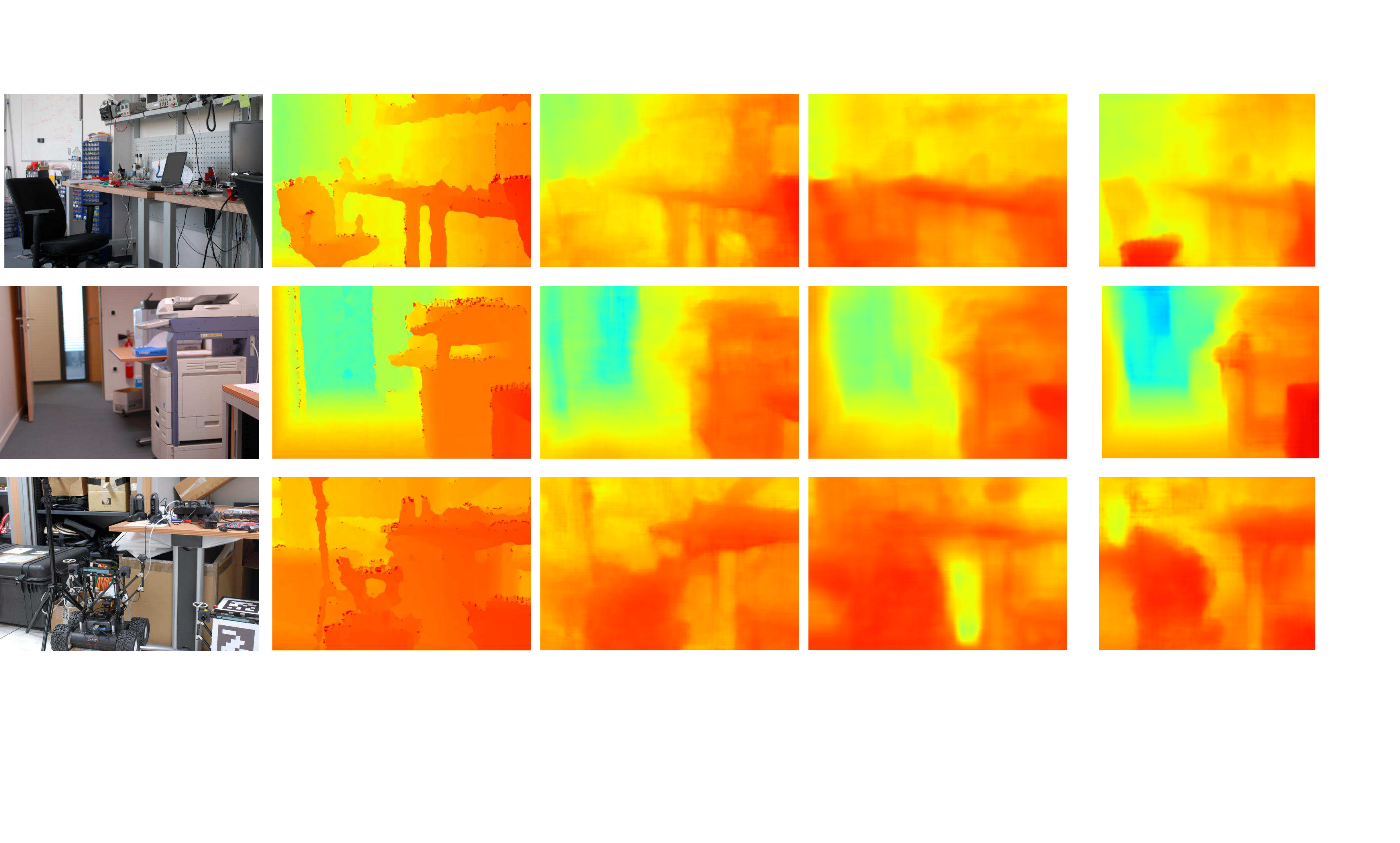}
  \caption{Qualitative comparison of D3-Net trained on defocused and all-focused images from a DSLR camera.}
  \label{fig:dfd_prediction}
  \end{figure}

\textbf{Performance results.} Using the real images dataset, we perform three experiments: first we train \dnet with the in-focus dataset and defocused dataset respectively, using same patch approach from last experiments. We also test \dnet with the in-focus dataset using an strategy that explores the global information of the scene using a series of preprocessing methods: we resize input images to 320x256 and performance data augmentation suggested in~\cite{Eigen2014} to improve generalization.

In Table~\ref{table:dfd_performance}, the performance metrics from the proposed models can be compared. The results show that defocus blur does improve the network performance increasing 10 to 20 percentual points in accuracy and also gives qualitative results with better segmentation as illustrated in Figure~\ref{fig:dfd_prediction}. 

  \begin{table}[]
    \caption{\label{table:dfd_performance} Performance comparison of D3-Net using all-in-focus and defocused images on a real DSLR dataset. }
    \centering
    \scalebox{1.0}{
      \begin{tabular}{p{2.0cm}@{\hspace{1.2mm}}c@{\hspace{1.2mm}}c@{\hspace{1.2mm}}c@{\hspace{0.5mm}}c@{}r@{}r@{\hspace{1mm}}r@{\hspace{1mm}}r@{\hspace{1mm}}r@{\hspace{1mm}}}\toprule
        {\multirow{2}{*}{Methods}} & \multicolumn{4}{c}{Error$\downarrow$} & \phantom{abc} &  \multicolumn{3}{c}{Accuracy$\uparrow$} \\
        \cmidrule{2-5} \cmidrule{7-9}
        & rel & log10 & rms & rmslog &&  \scalebox{.65}{$\delta\!<\!1.25$} & \scalebox{.65}{$\delta\!<\!1.25^2$} & \scalebox{.65}{$\delta\!<\!1.25^3$}\\  \midrule
        \fnum=2.8         & 0.157 & 0.065 & 0.546 & 0.234 && 80.9\% & 94.4\% & 97.6\% \\
        \fnum=8           & 0.225 & 0.095 & 0.730 & 0.285 && 60.2\% & 87.7\% & 98.0\% \\
        \fnum=8 (resize)  & 0.199 & 0.084 & 0.654 & 0.259 && 69.6\% & 91.6\% & 97.4\% \\ 
        \bottomrule
      \end{tabular}
      }
    \end{table}
The network is capable to find a relation between depth and defocus blur and predict better results, even thought the network may miss from global information when being trained with small patches. When feeding the network with resized images, filters from the last layers of the encoder, as from the first layers of the decoder, can understand the global information as they are fed with feature maps from the entire scene in a low resolution. However, this relation is not enough to give better predictions. As we can observe in the first examples of the third row in Figure~\ref{fig:dfd_prediction}, the DFD D3-Net used defocus to find the contours of the object, meanwhile the D3-Net with resize wrongly predicts the form of a chair, as it is an object constantly present in front of a desk. Our experiments show that the Deep-DFD model is more robust to generalization and less prone to overfitting than traditional methods trained and finetuned on all-in-focus images.

\section{Depth ``in the Wild''\label{sec:diw}}

In the era of autonomous driving vehicles (on land, on water, or in the air), there has been an increasing demand of less intrusive, more robust sensors and processing techniques to embed in systems able to evolve in the wild.
Previously, we validated our approach with several experiments on indoor scenes and we proved that blur can be learned by a neural network to improve prediction and also to improve the model's confidence to its estimations.
In this section, we now propose to tackle the general case of uncontrolled scenes.
We first assess the ability of the standard D3-Net, trained without defocus blur, to generalize to "in-the-wild" images using the Depth-in-the-Wild dataset~\cite{chen2016single} (DiW). Second, we use the whole system, D3-Net trained on indoor defocused images and the DSLR camera described from section~\ref{sec:real_dfd}, in uncontrolled, outdoor environments.

\begin{figure}[]
  \vspace{-3mm}
  \centering
    \def\svgwidth{1.0\textwidth}
    \scalebox{1.0}{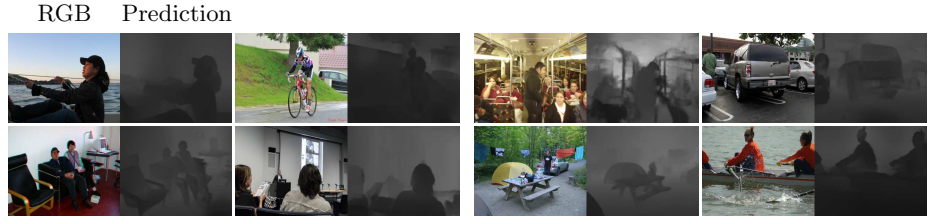}
    \caption{Examples of depth prediction using DIW dataset with D3-Net trained on \nyu.\label{fig:diw}}
    \vspace{-3mm}
\end{figure}

\textbf{Depth-in-the-Wild dataset (DiW).}
The ground truth of the DiW dataset is not dense; indeed, only two points of each RGB image are relatively annotated as being closer or farther from the camera, or at the same distance.
To adapt the network, we replace the objective function of D3-Net by the one proposed by the authors of the dataset~\cite{chen2016wild}. Then, for training, we take the weights of D3-Net trained on all-in-focus \nyu~\cite{Carvalho2018icip}, and finetune the model on DiW using the modified network.
We show the results of this model on the test set of DiW in figure~\ref{fig:diw}. The predicted depths present sharp edges for people and objects and give plausible estimates of the 3D structure of the given scenes. However, as the network was mostly trained on indoor scenes, it cannot give accurate depth predictions on sky regions. This shows that the a neural network has inherent capacity to predict depth in the wild. We will now see that we can improve this capacity by integrating physical cues of the sensor. 

\textbf{Deep-DFD in the wild.} We now propose to observe how deep models trained with blurred indoor images behave when confronted to challenging outdoor scenes. 
These experiments explore the model's capability to adapt predictions to new scenarios, never seen during training.
To perform our tests, we first acquire new data using the DSLR camera with defocus optics (from section~\ref{sec:real_dfd}) and keeping the same camera settings. As the depth sensor from the proposed platform works poorly outdoor, this new set of images does not contain respective depth ground truth. Thus, the model is neither trained on the new data, nor finetuned. Indeed, we use directly the models finetuned on indoor data with defocus blur (section~\ref{sec:real_dfd}).

\begin{figure*}[]
  \centering
  \def\svgwidth{1.0\textwidth}
  \scalebox{1.0}{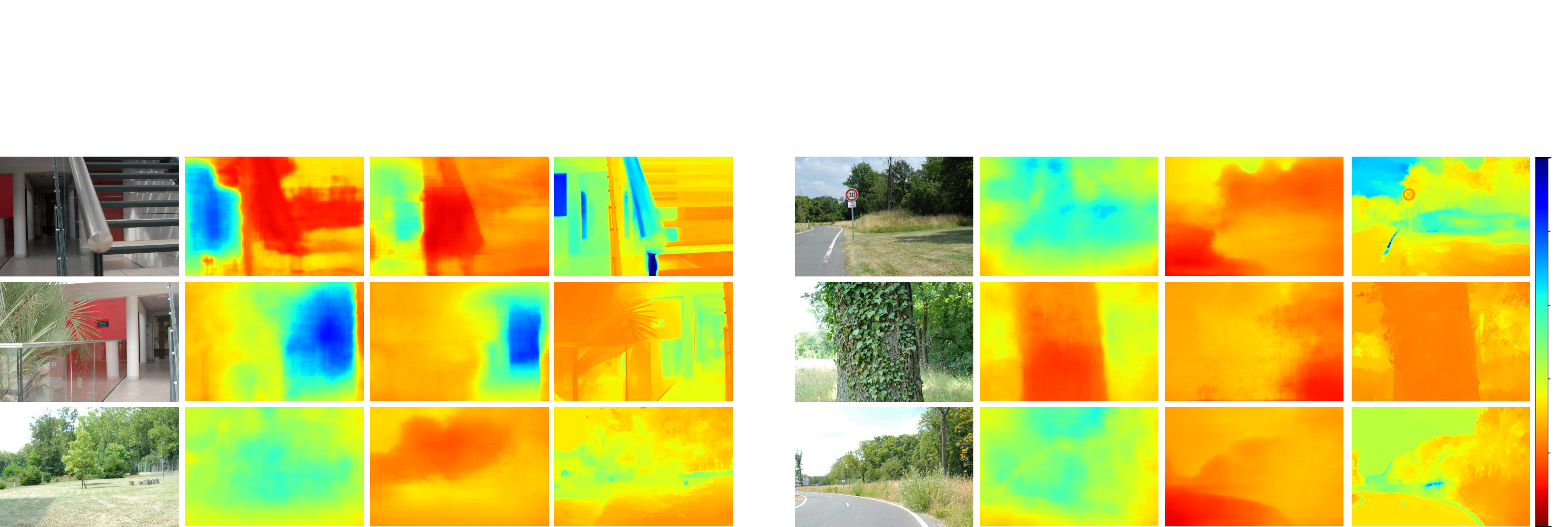}
  \caption{Comparison of monocular 3D estimation methods: from left to right, D3-Net trained on defocused images, D3-Net trained on all-in-focus images and a classical Depth from Defocus approach by~\cite{Zhuo11}. }
  \label{fig:deep_dfd_outdoor}
\end{figure*}

Results from the CNN models and from Zhuo's~\cite{Zhuo11} analytical method are shown in Figure~\ref{fig:deep_dfd_outdoor}. With D3-Net trained on all-in-focus images, the model constantly fails to extract information from new objects, as can be observed in the images with the road and also with the tree trunk. As expected, this model tries to base prediction on objects similar to what those seen during training or during finetuning, which are mostly non-existent in these new scenes. On the contrary, though the model trained with defocus blur information has equally never seen these new scenarios, the predictions give results relatively close to the expected depth maps. Indeed, the Deep-DFD model notably extracts and uses blur information to help prediction, as geometric features are unknown for the trained network. Finally, Zhuo's method also gives encouraging results, but constantly fails duo to defocus blur ambiguity to the focal plane (as on the handrail on the top left example of fig.~\ref{fig:deep_dfd_outdoor}). As can be deduced from our experiments, the combined use of geometric, statistical and defocus blur is a promising method to generalize learning capabilities.

\section{Conclusion}
In this paper, we have studied the influence of defocus blur as a cue in a monocular depth estimation using deep learning approach. We have shown that using blurred images outperforms the use of all-in-focus images, without requiring any scene model nor blur calibration. Besides, the combined use of defocus blur and geometrical structure information on the image, brought by the use of a deep network, avoids the classical limitations of Depth from Defocus with a conventional camera, such as depth ambiguity or dead zones. We have proposed different tools to visualize the benefit of defocus blur on the network performance, such as per depth error statistics and uncertainty maps. These tools have shown that depth estimation with defocus blur is most significantly improved at short depths, resulting in better depth map segmentations. 
We have also compared performance of deep depth estimation with defocus blur from several optical settings to better understand the influence of the camera parameters to deep depth prediction. In our tests, the best performances are obtained for a close in-focus plane, which leads to really small camera depths of field and thus defocus blur on most of the objects in the dataset.

Besides synthetic data, this paper also provides excellent results on both indoor and outdoor real defocused images from a new set of DSLR images.  
These experiments on real defocused data proved that defocus blur combined to neural networks are more robust to training data and domain generalization, reducing possible constraints of actual acquisition models with active sensors and stereo systems. Notably, results on the challenging domain of outdoor scenes without further calibration, or finetuning prove that this new system can be widely used in the wild to combine physical information (defocus blur) and cues already used by standard neural networks, such as geometry and perspective.
These observations open the way to further studies on the optimization of the camera parameters and acquisition modalities for 3D estimation using defocus blur and deep learning.

\clearpage

\bibliographystyle{splncs}
\bibliography{biblio_depth}

\end{document}

%% file: images/visual_abstract.pdf_tex
\begingroup%
  \makeatletter%
  \providecommand\color[2][]{%
    \errmessage{(Inkscape) Color is used for the text in Inkscape, but the package 'color.sty' is not loaded}%
    \renewcommand\color[2][]{}%
  }%
  \providecommand\transparent[1]{%
    \errmessage{(Inkscape) Transparency is used (non-zero) for the text in Inkscape, but the package 'transparent.sty' is not loaded}%
    \renewcommand\transparent[1]{}%
  }%
  \providecommand\rotatebox[2]{#2}%
  \ifx\svgwidth\undefined%
    \setlength{\unitlength}{549.40315494bp}%
    \ifx\svgscale\undefined%
      \relax%
    \else%
      \setlength{\unitlength}{\unitlength * \real{\svgscale}}%
    \fi%
  \else%
    \setlength{\unitlength}{\svgwidth}%
  \fi%
  \global\let\svgwidth\undefined%
  \global\let\svgscale\undefined%
  \makeatother%
  \begin{picture}(1,0.46381744)%
    \put(0,0){\includegraphics[width=\unitlength]{visual_abstract.pdf}}%
    \put(0.75629083,0.18588376){\color[rgb]{0,0,0}\makebox(0,0)[b]{\smash{Outoor scene with \textbf{real} defocus}}}%
    \put(0.75315906,0.06071348){\color[rgb]{0,0,0}\makebox(0,0)[b]{\smash{CNN}}}%
    \put(0.75629083,0.02120106){\color[rgb]{0,0,0}\makebox(0,0)[b]{\smash{\textcolor{myblue}{Transfer learning}}}}%
    \put(0.24417032,0.18737414){\color[rgb]{0,0,0}\makebox(0,0)[b]{\smash{Indoor  scene with \textbf{real} defocus}}}%
    \put(0.24135106,0.07320188){\color[rgb]{0,0,0}\makebox(0,0)[b]{\smash{CNN}}}%
    \put(0.24417032,0.0184266){\color[rgb]{0,0,0}\makebox(0,0)[b]{\smash{\textcolor{myblue}{Transfer learning + finetuning}}}}%
    \put(0.4966853,0.31811934){\color[rgb]{0,0,0}\makebox(0,0)[b]{\smash{CNN}}}%
    \put(0.49950464,0.43229162){\color[rgb]{0,0,0}\makebox(0,0)[b]{\smash{Indoor scene with \textbf{synthetic} defocus}}}%
    \put(0.49950464,0.26334409){\color[rgb]{0,0,0}\makebox(0,0)[b]{\smash{\textcolor{myblue}{Supervised learning}}}}%
  \end{picture}%
\endgroup%

%% file: images/nyuv2_blur_extracts.pdf_tex
\begingroup%
  \makeatletter%
  \providecommand\color[2][]{%
    \errmessage{(Inkscape) Color is used for the text in Inkscape, but the package 'color.sty' is not loaded}%
    \renewcommand\color[2][]{}%
  }%
  \providecommand\transparent[1]{%
    \errmessage{(Inkscape) Transparency is used (non-zero) for the text in Inkscape, but the package 'transparent.sty' is not loaded}%
    \renewcommand\transparent[1]{}%
  }%
  \providecommand\rotatebox[2]{#2}%
  \ifx\svgwidth\undefined%
    \setlength{\unitlength}{3552.946bp}%
    \ifx\svgscale\undefined%
      \relax%
    \else%
      \setlength{\unitlength}{\unitlength * \real{\svgscale}}%
    \fi%
  \else%
    \setlength{\unitlength}{\svgwidth}%
  \fi%
  \global\let\svgwidth\undefined%
  \global\let\svgscale\undefined%
  \makeatother%
  \begin{picture}(1,0.16955788)%
    \put(0,0){\includegraphics[width=\unitlength,page=1]{nyuv2_blur_extracts.pdf}}%
  \end{picture}%
\endgroup%

%% file: images/nyu_prediction.pdf_tex
\begingroup%
  \makeatletter%
  \providecommand\color[2][]{%
    \errmessage{(Inkscape) Color is used for the text in Inkscape, but the package 'color.sty' is not loaded}%
    \renewcommand\color[2][]{}%
  }%
  \providecommand\transparent[1]{%
    \errmessage{(Inkscape) Transparency is used (non-zero) for the text in Inkscape, but the package 'transparent.sty' is not loaded}%
    \renewcommand\transparent[1]{}%
  }%
  \providecommand\rotatebox[2]{#2}%
  \ifx\svgwidth\undefined%
    \setlength{\unitlength}{289.02749309bp}%
    \ifx\svgscale\undefined%
      \relax%
    \else%
      \setlength{\unitlength}{\unitlength * \real{\svgscale}}%
    \fi%
  \else%
    \setlength{\unitlength}{\svgwidth}%
  \fi%
  \global\let\svgwidth\undefined%
  \global\let\svgscale\undefined%
  \makeatother%
  \begin{picture}(1,0.61290868)%
    \put(0,0){\includegraphics[width=\unitlength]{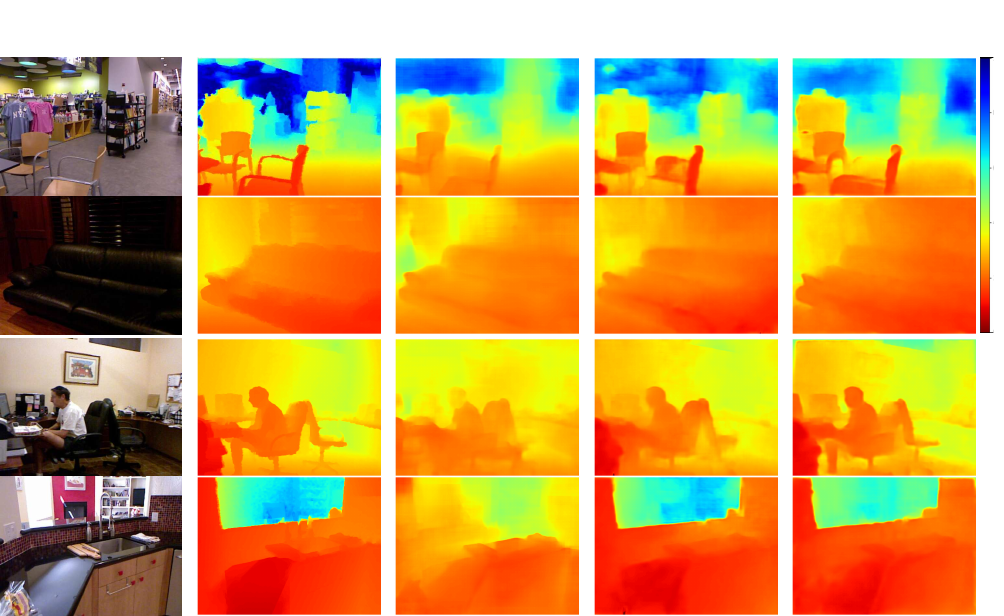}}%
    \put(0.88036117,0.6018912){\color[rgb]{0,0,0}\makebox(0,0)[b]{\smash{D3-Net}}}%
    \put(0.68282935,0.6018912){\color[rgb]{0,0,0}\makebox(0,0)[b]{\smash{D3-Net}}}%
    \put(0.28834186,0.5739961){\color[rgb]{0,0,0}\makebox(0,0)[b]{\smash{Truth}}}%
    \put(0.09023307,0.5739961){\color[rgb]{0,0,0}\makebox(0,0)[b]{\smash{all-in-focus}}}%
    \put(0.48529724,0.5739961){\color[rgb]{0,0,0}\makebox(0,0)[b]{\smash{All-in-focus}}}%
    \put(0.09023307,0.6018912){\color[rgb]{0,0,0}\makebox(0,0)[b]{\smash{RGB}}}%
    \put(0.28834186,0.6017623){\color[rgb]{0,0,0}\makebox(0,0)[b]{\smash{Ground}}}%
    \put(0.68262721,0.5738922){\color[rgb]{0,0,0}\makebox(0,0)[b]{\smash{focus at 2m}}}%
    \put(0.87983729,0.5739961){\color[rgb]{0,0,0}\makebox(0,0)[b]{\smash{focus at 8m}}}%
    \put(0.99008365,0.2820419){\color[rgb]{0,0,0}\makebox(0,0)[lb]{\smash{\tiny{0m}}}}%
    \put(0.99005447,0.3354732){\color[rgb]{0,0,0}\makebox(0,0)[lb]{\smash{\tiny{2}}}}%
    \put(0.99015177,0.39136679){\color[rgb]{0,0,0}\makebox(0,0)[lb]{\smash{\tiny{4}}}}%
    \put(0.99006809,0.44559189){\color[rgb]{0,0,0}\makebox(0,0)[lb]{\smash{\tiny{6}}}}%
    \put(0.99007585,0.50065267){\color[rgb]{0,0,0}\makebox(0,0)[lb]{\smash{\tiny{8}}}}%
    \put(0.98990849,0.55529774){\color[rgb]{0,0,0}\makebox(0,0)[lb]{\smash{\tiny{10}}}}%
  \end{picture}%
\endgroup%

%% file: images/nyu_blur_comparison.pdf_tex
\begingroup%
  \makeatletter%
  \providecommand\color[2][]{%
    \errmessage{(Inkscape) Color is used for the text in Inkscape, but the package 'color.sty' is not loaded}%
    \renewcommand\color[2][]{}%
  }%
  \providecommand\transparent[1]{%
    \errmessage{(Inkscape) Transparency is used (non-zero) for the text in Inkscape, but the package 'transparent.sty' is not loaded}%
    \renewcommand\transparent[1]{}%
  }%
  \providecommand\rotatebox[2]{#2}%
  \ifx\svgwidth\undefined%
    \setlength{\unitlength}{571.90713746bp}%
    \ifx\svgscale\undefined%
      \relax%
    \else%
      \setlength{\unitlength}{\unitlength * \real{\svgscale}}%
    \fi%
  \else%
    \setlength{\unitlength}{\svgwidth}%
  \fi%
  \global\let\svgwidth\undefined%
  \global\let\svgscale\undefined%
  \makeatother%
  \begin{picture}(1,0.2329954)%
    \put(0,0){\includegraphics[width=\unitlength]{nyu_blur_comparison.pdf}}%
    \put(0.26686124,0.20594466){\color[rgb]{0,0,0}\makebox(0,0)[lt]{\begin{minipage}{0.18600955\unitlength}\centering Ground truth\end{minipage}}}%
    \put(0.51550761,0.20594466){\color[rgb]{0,0,0}\makebox(0,0)[lt]{\begin{minipage}{0.18600955\unitlength}\centering Zhuo~\cite{Zhuo11} \end{minipage}}}%
    \put(0.76415398,0.20594466){\color[rgb]{0,0,0}\makebox(0,0)[lt]{\begin{minipage}{0.18600955\unitlength}\centering D3-Net\end{minipage}}}%
    \put(0.0186063,0.20806421){\color[rgb]{0,0,0}\makebox(0,0)[lt]{\begin{minipage}{0.18600955\unitlength}\centering focus at 2m\end{minipage}}}%
    \put(0.0186063,0.23703133){\color[rgb]{0,0,0}\makebox(0,0)[lt]{\begin{minipage}{0.18600955\unitlength}\centering RGB\end{minipage}}}%
    \put(0.99389257,0.00241795){\color[rgb]{0,0,0}\makebox(0,0)[lb]{\smash{\tiny{0m}}}}%
    \put(0.99387459,0.03532603){\color[rgb]{0,0,0}\makebox(0,0)[lb]{\smash{\tiny{2}}}}%
    \put(0.99393452,0.06975062){\color[rgb]{0,0,0}\makebox(0,0)[lb]{\smash{\tiny{4}}}}%
    \put(0.99388298,0.1031476){\color[rgb]{0,0,0}\makebox(0,0)[lb]{\smash{\tiny{6}}}}%
    \put(0.99388776,0.13705927){\color[rgb]{0,0,0}\makebox(0,0)[lb]{\smash{\tiny{8}}}}%
    \put(0.99378469,0.1707149){\color[rgb]{0,0,0}\makebox(0,0)[lb]{\smash{\tiny{10}}}}%
  \end{picture}%
\endgroup%

%% file: images/nyuv2_uncertainty.pdf_tex
\begingroup%
  \makeatletter%
  \providecommand\color[2][]{%
    \errmessage{(Inkscape) Color is used for the text in Inkscape, but the package 'color.sty' is not loaded}%
    \renewcommand\color[2][]{}%
  }%
  \providecommand\transparent[1]{%
    \errmessage{(Inkscape) Transparency is used (non-zero) for the text in Inkscape, but the package 'transparent.sty' is not loaded}%
    \renewcommand\transparent[1]{}%
  }%
  \providecommand\rotatebox[2]{#2}%
  \ifx\svgwidth\undefined%
    \setlength{\unitlength}{3129.68810406bp}%
    \ifx\svgscale\undefined%
      \relax%
    \else%
      \setlength{\unitlength}{\unitlength * \real{\svgscale}}%
    \fi%
  \else%
    \setlength{\unitlength}{\svgwidth}%
  \fi%
  \global\let\svgwidth\undefined%
  \global\let\svgscale\undefined%
  \makeatother%
  \begin{picture}(1,0.31982628)%
    \put(0,0){\includegraphics[width=\unitlength]{nyuv2_uncertainty.pdf}}%
    \put(0.05771651,0.30491787){\color[rgb]{0,0,0}\makebox(0,0)[b]{\smash{RGB}}}%
    \put(0.17765578,0.304828){\color[rgb]{0,0,0}\makebox(0,0)[b]{\smash{Ground Truth}}}%
    \put(0.54749991,0.31205793){\color[rgb]{0,0,0}\makebox(0,0)[b]{\smash{All in focus}}}%
    \put(0.54715792,0.29672094){\color[rgb]{0,0,0}\makebox(0,0)[b]{\smash{epistemic uncertainty}}}%
    \put(0.30824161,0.31131141){\color[rgb]{0,0,0}\makebox(0,0)[b]{\smash{All in focus}}}%
    \put(0.30800447,0.29672094){\color[rgb]{0,0,0}\makebox(0,0)[b]{\smash{prediction}}}%
    \put(0.42787075,0.31205793){\color[rgb]{0,0,0}\makebox(0,0)[b]{\smash{All in focus}}}%
    \put(0.42719926,0.29675089){\color[rgb]{0,0,0}\makebox(0,0)[b]{\smash{mean error}}}%
    \put(0.91591327,0.29672093){\color[rgb]{0,0,0}\makebox(0,0)[b]{\smash{epistemic uncertainty}}}%
    \put(0.91579345,0.31205792){\color[rgb]{0,0,0}\makebox(0,0)[b]{\smash{DFD focus at 2m}}}%
    \put(0.67751211,0.31205793){\color[rgb]{0,0,0}\makebox(0,0)[b]{\smash{DFD focus at 2m}}}%
    \put(0.67773677,0.29672094){\color[rgb]{0,0,0}\makebox(0,0)[b]{\smash{prediction}}}%
    \put(0.79665275,0.31205793){\color[rgb]{0,0,0}\makebox(0,0)[b]{\smash{DFD focus at 2m}}}%
    \put(0.79644306,0.29672094){\color[rgb]{0,0,0}\makebox(0,0)[b]{\smash{mean error}}}%
    \put(0.99144031,0.02151796){\color[rgb]{0,0,0}\makebox(0,0)[lb]{\smash{\tiny{0}}}}%
    \put(0.99141277,0.07194188){\color[rgb]{0,0,0}\makebox(0,0)[lb]{\smash{\tiny{2}}}}%
    \put(0.99150459,0.12468948){\color[rgb]{0,0,0}\makebox(0,0)[lb]{\smash{\tiny{4}}}}%
    \put(0.99142563,0.17586252){\color[rgb]{0,0,0}\makebox(0,0)[lb]{\smash{\tiny{6}}}}%
    \put(0.99143295,0.2278242){\color[rgb]{0,0,0}\makebox(0,0)[lb]{\smash{\tiny{8}}}}%
    \put(0.99127501,0.27939356){\color[rgb]{0,0,0}\makebox(0,0)[lb]{\smash{\tiny{10}}}}%
    \put(0.95757838,0.00154816){\color[rgb]{0,0,0}\makebox(0,0)[b]{\smash{depth (*1m)}}}%
    \put(0.89656735,0.00132102){\color[rgb]{0,0,0}\makebox(0,0)[rb]{\smash{variance (*10m$^2$)}}}%
  \end{picture}%
\endgroup%

%% file: images/dfd_prediction.pdf_tex
\begingroup%
  \makeatletter%
  \providecommand\color[2][]{%
    \errmessage{(Inkscape) Color is used for the text in Inkscape, but the package 'color.sty' is not loaded}%
    \renewcommand\color[2][]{}%
  }%
  \providecommand\transparent[1]{%
    \errmessage{(Inkscape) Transparency is used (non-zero) for the text in Inkscape, but the package 'transparent.sty' is not loaded}%
    \renewcommand\transparent[1]{}%
  }%
  \providecommand\rotatebox[2]{#2}%
  \ifx\svgwidth\undefined%
    \setlength{\unitlength}{2790.79473469bp}%
    \ifx\svgscale\undefined%
      \relax%
    \else%
      \setlength{\unitlength}{\unitlength * \real{\svgscale}}%
    \fi%
  \else%
    \setlength{\unitlength}{\svgwidth}%
  \fi%
  \global\let\svgwidth\undefined%
  \global\let\svgscale\undefined%
  \makeatother%
  \begin{picture}(1,0.60236696)%
    \put(0.86216645,0.56148727){\color[rgb]{0,0,0}\makebox(0,0)[b]{\smash{N=8 (resize)}}}%
    \put(0.66979318,0.56091339){\color[rgb]{0,0,0}\makebox(0,0)[b]{\smash{N=8}}}%
    \put(0.28721771,0.56081261){\color[rgb]{0,0,0}\makebox(0,0)[b]{\smash{Truth}}}%
    \put(0.09530992,0.56081261){\color[rgb]{0,0,0}\makebox(0,0)[b]{\smash{all-in-focus}}}%
    \put(0.47833607,0.56091339){\color[rgb]{0,0,0}\makebox(0,0)[b]{\smash{DFD N=2.8}}}%
    \put(0,0){\includegraphics[width=\unitlength,page=1]{dfd_prediction.pdf}}%
    \put(0.09530992,0.59375606){\color[rgb]{0,0,0}\makebox(0,0)[b]{\smash{RGB}}}%
    \put(0.47833607,0.59375606){\color[rgb]{0,0,0}\makebox(0,0)[b]{\smash{DSLR}}}%
    \put(0.66956643,0.59375606){\color[rgb]{0,0,0}\makebox(0,0)[b]{\smash{DSLR}}}%
    \put(0.86183892,0.59375606){\color[rgb]{0,0,0}\makebox(0,0)[b]{\smash{DSLR}}}%
    \put(0.28721771,0.59365528){\color[rgb]{0,0,0}\makebox(0,0)[b]{\smash{Ground}}}%
    \put(0,0){\includegraphics[width=\unitlength,page=2]{dfd_prediction.pdf}}%
    \put(0.98568536,0.13870024){\color[rgb]{0,0,0}\makebox(0,0)[lb]{\smash{\tiny{0m}}}}%
    \put(0.98564323,0.21583043){\color[rgb]{0,0,0}\makebox(0,0)[lb]{\smash{\tiny{2}}}}%
    \put(0.98578369,0.29651502){\color[rgb]{0,0,0}\makebox(0,0)[lb]{\smash{\tiny{4}}}}%
    \put(0.9856629,0.37479109){\color[rgb]{0,0,0}\makebox(0,0)[lb]{\smash{\tiny{6}}}}%
    \put(0.98567409,0.4542735){\color[rgb]{0,0,0}\makebox(0,0)[lb]{\smash{\tiny{8}}}}%
    \put(0.98543251,0.53315581){\color[rgb]{0,0,0}\makebox(0,0)[lb]{\smash{\tiny{10}}}}%
  \end{picture}%
\endgroup%

%% file: images/diw_results.pdf_tex
\begingroup%
  \makeatletter%
  \providecommand\color[2][]{%
    \errmessage{(Inkscape) Color is used for the text in Inkscape, but the package 'color.sty' is not loaded}%
    \renewcommand\color[2][]{}%
  }%
  \providecommand\transparent[1]{%
    \errmessage{(Inkscape) Transparency is used (non-zero) for the text in Inkscape, but the package 'transparent.sty' is not loaded}%
    \renewcommand\transparent[1]{}%
  }%
  \providecommand\rotatebox[2]{#2}%
  \ifx\svgwidth\undefined%
    \setlength{\unitlength}{2103.41762695bp}%
    \ifx\svgscale\undefined%
      \relax%
    \else%
      \setlength{\unitlength}{\unitlength * \real{\svgscale}}%
    \fi%
  \else%
    \setlength{\unitlength}{\svgwidth}%
  \fi%
  \global\let\svgwidth\undefined%
  \global\let\svgscale\undefined%
  \makeatother%
  \begin{picture}(1,0.21953973)%
    \put(0,0){\includegraphics[width=\unitlength]{diw_results.pdf}}%
    \put(0.0608364,0.21224668){\color[rgb]{0,0,0}\makebox(0,0)[b]{\smash{RGB}}}%
    \put(0.18356382,0.21224668){\color[rgb]{0,0,0}\makebox(0,0)[b]{\smash{Prediction}}}%
  \end{picture}%
\endgroup%

%% file: images/dfd_outdoor_prediction_2.pdf_tex
\begingroup%
  \makeatletter%
  \providecommand\color[2][]{%
    \errmessage{(Inkscape) Color is used for the text in Inkscape, but the package 'color.sty' is not loaded}%
    \renewcommand\color[2][]{}%
  }%
  \providecommand\transparent[1]{%
    \errmessage{(Inkscape) Transparency is used (non-zero) for the text in Inkscape, but the package 'transparent.sty' is not loaded}%
    \renewcommand\transparent[1]{}%
  }%
  \providecommand\rotatebox[2]{#2}%
  \ifx\svgwidth\undefined%
    \setlength{\unitlength}{4527.42880063bp}%
    \ifx\svgscale\undefined%
      \relax%
    \else%
      \setlength{\unitlength}{\unitlength * \real{\svgscale}}%
    \fi%
  \else%
    \setlength{\unitlength}{\svgwidth}%
  \fi%
  \global\let\svgwidth\undefined%
  \global\let\svgscale\undefined%
  \makeatother%
  \begin{picture}(1,0.33587278)%
    \put(0,0){\includegraphics[width=\unitlength]{dfd_outdoor_prediction_2.pdf}}%
    \put(0.99117618,0.00077023){\color[rgb]{0,0,0}\makebox(0,0)[lb]{\smash{\tiny{0m}}}}%
    \put(0.99115021,0.04665442){\color[rgb]{0,0,0}\makebox(0,0)[lb]{\smash{\tiny{2}}}}%
    \put(0.99123679,0.0946531){\color[rgb]{0,0,0}\makebox(0,0)[lb]{\smash{\tiny{4}}}}%
    \put(0.99116233,0.14121896){\color[rgb]{0,0,0}\makebox(0,0)[lb]{\smash{\tiny{6}}}}%
    \put(0.99116923,0.18850247){\color[rgb]{0,0,0}\makebox(0,0)[lb]{\smash{\tiny{8}}}}%
    \put(0.99102032,0.23542898){\color[rgb]{0,0,0}\makebox(0,0)[lb]{\smash{\tiny{10}}}}%
    \put(0.35441584,0.30263069){\color[rgb]{0,0,0}\makebox(0,0)[lt]{\begin{minipage}{0.11209468\unitlength}\centering Zhuo~\cite{Zhuo11}\\ N=2.8 \\ \end{minipage}}}%
    \put(0.23681714,0.33767078){\color[rgb]{0,0,0}\makebox(0,0)[lt]{\begin{minipage}{0.11209468\unitlength}\centering D3-Net\\ N=8\\ (resize) \\ \end{minipage}}}%
    \put(0.11881826,0.30264277){\color[rgb]{0,0,0}\makebox(0,0)[lt]{\begin{minipage}{0.11209468\unitlength}\centering D3-Net\\ N=2.8 \\ \end{minipage}}}%
    \put(0.0008004,0.30264277){\color[rgb]{0,0,0}\makebox(0,0)[lt]{\begin{minipage}{0.11209468\unitlength}\centering RGB\end{minipage}}}%
  \end{picture}%
\endgroup%